\documentclass[letterpaper]{article}
\providecommand\NewCommandCopy[2]{\let#1=#2}
\usepackage[preprint]{aaai2027}
\usepackage[hyphens]{url}
\usepackage{graphicx}
\usepackage{cuted}
\usepackage{natbib}
\usepackage{caption}
\usepackage{booktabs}
\usepackage{colortbl}
\usepackage{amsmath}
\usepackage{amssymb}
\usepackage{algorithm}
\usepackage{algorithmic}
\usepackage{float}
\definecolor{pneccol}{gray}{0.93}
\definecolor{gaincolor}{RGB}{0,128,0}
\newcommand{\gain}[2]{%
  \textbf{#1}\,{\color{gaincolor}$(+#2)$}%
}
\definecolor{revisionblue}{RGB}{0,0,255}

\newcommand{\res}[2]{\shortstack{$#1$\\$\pm #2$}}
\newcommand{\bestres}[2]{\shortstack{$\mathbf{#1}$\\$\mathbf{\pm #2}$}}
\newcommand{\tablefont}{\fontsize{9}{10}\selectfont}

\title{PNEC-Mamba: Prototype-Guided Positive-Negative Evidence Calibration for Hyperspectral Image Classification}

\author{
	Mingzhen Xu,
	Can Xu,
	Di Wang,
	Haonan Guo,
	Bo Du
}
\affiliations{}

\begin{document}
\maketitle

\begin{abstract}
In real-world hyperspectral scenes, pixel representations are often ambiguous due to factors such as  spectral similarity, mixed pixels, and local context interference, which may simultaneously encode discriminative evidence and interfering information. Existing methods mainly focus on learning more powerful representations or modeling broader contexts, but rarely investigate whether the learned representations provide reliable evidence or introduce interference into classification decisions. To address this issue, we view hyperspectral image classification from the perspective of pixel-level evidence reliability modeling and propose PNEC-Mamba, a prototype-guided positive–negative evidence calibration framework. The framework progressively establishes semantic references, separates class-related evidence from interference, estimates pixel-level reliability, and performs selective calibration. First, a full-image state-space encoder extracts pixel representations, while dynamic class prototypes provide semantic references that evolve jointly with the feature space. Subsequently, positive and negative evidence is derived from pixel–prototype competition, explicitly separating discriminative cues that support classification from confusing signals associated with competing classes. Based on these evidence relationships, a multi-source uncertainty estimation strategy is introduced to assess pixel-level reliability, enabling stronger evidence calibration for uncertain regions. Finally, a full-resolution consistency refinement step is applied to recover local spatial details and improve boundary coherence in the final predictions. Extensive experiments on three benchmark datasets demonstrate that PNEC-Mamba achieves superior classification performance compared with state-of-the-art methods.
\end{abstract}

\section{Introduction}

\begin{figure}[!t]
\centering
\includegraphics[width=0.9\columnwidth]{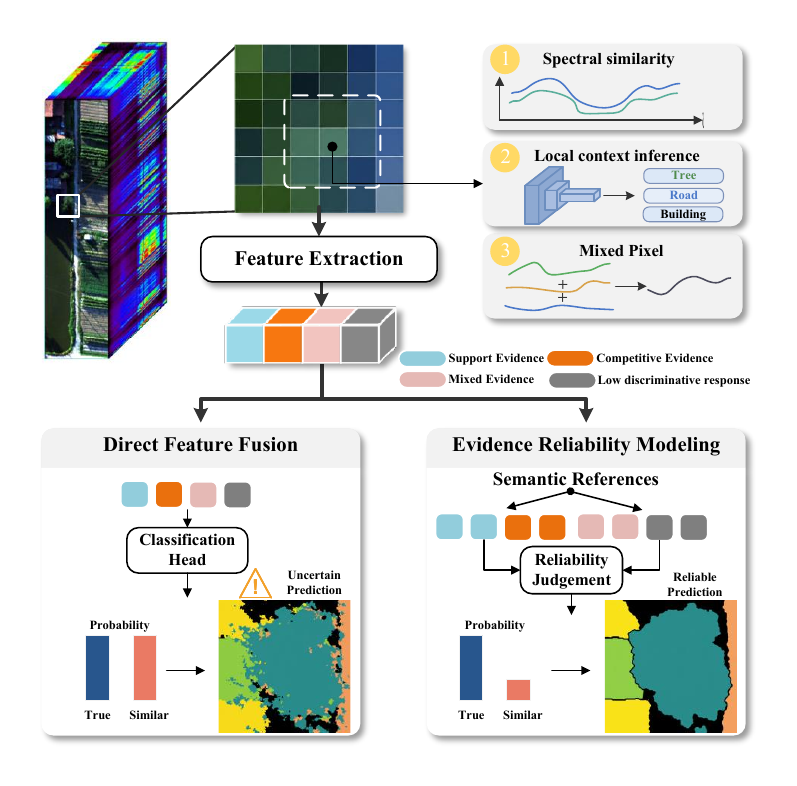}
\caption{Motivation for evidence reliability modeling. Spectral similarity, local context interference, and mixed pixels entangle supportive and competitive evidence, motivating reliability-aware evidence calibration.}
\label{fig:motivation}
\end{figure}

Hyperspectral imagery captures rich spectral signatures at each pixel, providing valuable spectral–spatial information for applications such as urban land-cover mapping, precision agriculture, environmental monitoring, and disaster assessment \citep{li2019deepoverview,zhu2026knowledge}. Assigning semantic labels to these pixels is a fundamental task in remote sensing image interpretation \citep{bioucas2013hyperspectral}, yet learning discriminative representations from high-dimensional spectra and complex scenes remains challenging. Deep learning methods have significantly improved this task by learning discriminative spectral–spatial representations directly from data \citep{song2025review}. CNN-based methods effectively capture local discriminative patterns \citep{hu2015deep}, but their restricted receptive fields limit long-range dependency modeling. Transformer-based methods alleviate this limitation through self-attention mechanisms \citep{vaswani2017attention,hong2022spectralformer}, but their quadratic complexity leads to substantial computational costs. More recently, Mamba-based selective state-space models have provided a linear-complexity alternative for long-range dependency modeling \citep{gu2023mamba}. Subsequent studies have further improved Mamba-based hyperspectral classification through enhanced spectral–spatial modeling and adaptive architectures \citep{yao2024spectralmamba,pan2025mambalg,xu2026mambamoe}. Nevertheless, these methods primarily focus on feature representation and fusion, while implicitly assuming that the resulting representations contain reliable information for classification.

In practice, a pixel representation may contain both positive evidence associated with class-consistent evidence and conflicting information caused by spectral similarity, mixed components, or neighborhood interference. Their entanglement increases prediction uncertainty and obscures which cues genuinely support the classification decision. Directly aggregating such representations may simultaneously reinforce discriminative and conflicting responses, leading to errors near class boundaries and in heterogeneous or highly mixed regions, as shown in Figure 1. Therefore, hyperspectral classification requires not only effective representation learning but also an explicit mechanism for distinguishing reliable discriminative evidence from confounding responses. Recent studies have improved the discrimination of ambiguous pixels through class-relation modeling and uncertainty-aware representation learning \citep{xu2026mambamoe,guo2026pcrnet,chu2026fuzzygraph,chu2026fuzzyhyperbolic}. Although these approaches improve the discrimination of ambiguous pixels, they mainly refine feature spaces, sample relationships, or learning objectives. The distinction between reliable evidence and conflicting interference remains insufficiently explored.

Motivated by these observations, we formulate hyperspectral image classification from the perspective of pixel-level evidence reliability modeling. Unlike methods that focus solely on feature representation learning, this formulation emphasizes the separation and adaptive calibration of classification evidence. Specifically, under class similarity, mixed pixels, and contextual interference, the model should identify which components of a pixel representation provide reliable discriminative evidence and which components introduce conflicting information from confusing classes or surrounding backgrounds. Based on this distinction, uncertain regions can be selectively calibrated without disrupting reliable predictions.

To this end, we propose PNEC-Mamba, a prototype-guided positive--negative evidence calibration framework for hyperspectral image classification. Specifically, PNEC-Mamba integrates a full-image spectral–spatial state-space encoder with a dynamic class prototype bank to establish adaptive semantic references in the evolving feature space. By exploiting pixel–prototype competition, the model derives positive and negative evidence representations, where positive evidence captures class-consistent cues and negative evidence characterizes responses associated with confusing classes. A multi-source uncertainty estimation mechanism is further introduced to evaluate pixel-level reliability and guide selective calibration. Finally, a full-resolution consistency refinement module is adopted to recover spatial details and improve boundary coherence after calibration. The main contributions of this work are summarized as follows:
\begin{itemize}
    \item We establish a pixel-level evidence reliability framework for hyperspectral image classification, extending conventional feature representation learning by explicitly modeling evidence construction, reliability estimation, and selective calibration.
    \item We develop a dynamic prototype-guided mechanism that captures the relationships between pixels and class prototypes, enabling positive evidence enhancement and negative evidence suppression.
    \item We introduce a reliability-aware calibration mechanism that uses uncertainty estimation and consistency refinement to improve ambiguous pixels while preserving reliable predictions and spatial coherence.
    \item Extensive experiments on three benchmark datasets demonstrate the effectiveness of PNEC-Mamba, with ablation studies and mechanism analyses further validating the contribution of each component.
\end{itemize}

\section{Related Work}

Deep learning has become the dominant paradigm for hyperspectral image classification by enabling effective spectral--spatial representation learning. CNN-based methods learn local discriminative patterns through convolution operations \citep{hu2015deep,roy2020hybridsn}, whereas Transformer-based approaches capture long-range dependencies through self-attention mechanisms \citep{hong2022spectralformer,sun2024massformer}. However, CNNs are constrained by limited receptive fields, and Transformers introduce substantial computational costs due to quadratic attention complexity. Recently, Mamba-based state-space models have attracted increasing attention because of their linear-complexity sequence modeling capability \citep{gu2023mamba}. Existing Mamba-based approaches improve hyperspectral classification through enhanced spectral modeling, spatial interaction, multiscale representation, and adaptive architectures \citep{yao2024spectralmamba,huang2024ssmamba,li2024mambahsi,wang2025s2mamba,pan2025mambalg,xu2026mambamoe}. Nevertheless, these methods primarily focus on improving feature representation and fusion, while the reliability of the learned representations remains largely unexplored.

Beyond representation learning, recent studies have explored class-relation modeling and uncertainty-aware strategies to improve the robustness of hyperspectral classification in complex scenes. Prototype-based methods represent categories through compact class references and refine pixel--class relationships \citep{snell2017prototypical,guo2026pcrnet}, while uncertainty-aware approaches model ambiguous samples through relational or probabilistic structures \citep{chu2026fuzzygraph,chu2026fuzzyhyperbolic}. These methods improve class discrimination and uncertainty handling; however, they mainly focus on refining feature relationships or prediction confidence rather than explicitly analyzing the evidence that contributes to each decision. Consequently, reliable evidence and conflicting information may still remain entangled at the pixel level.

\section{Methodology}

\begin{figure*}[t]
\centering
\includegraphics[width=0.78\textwidth]{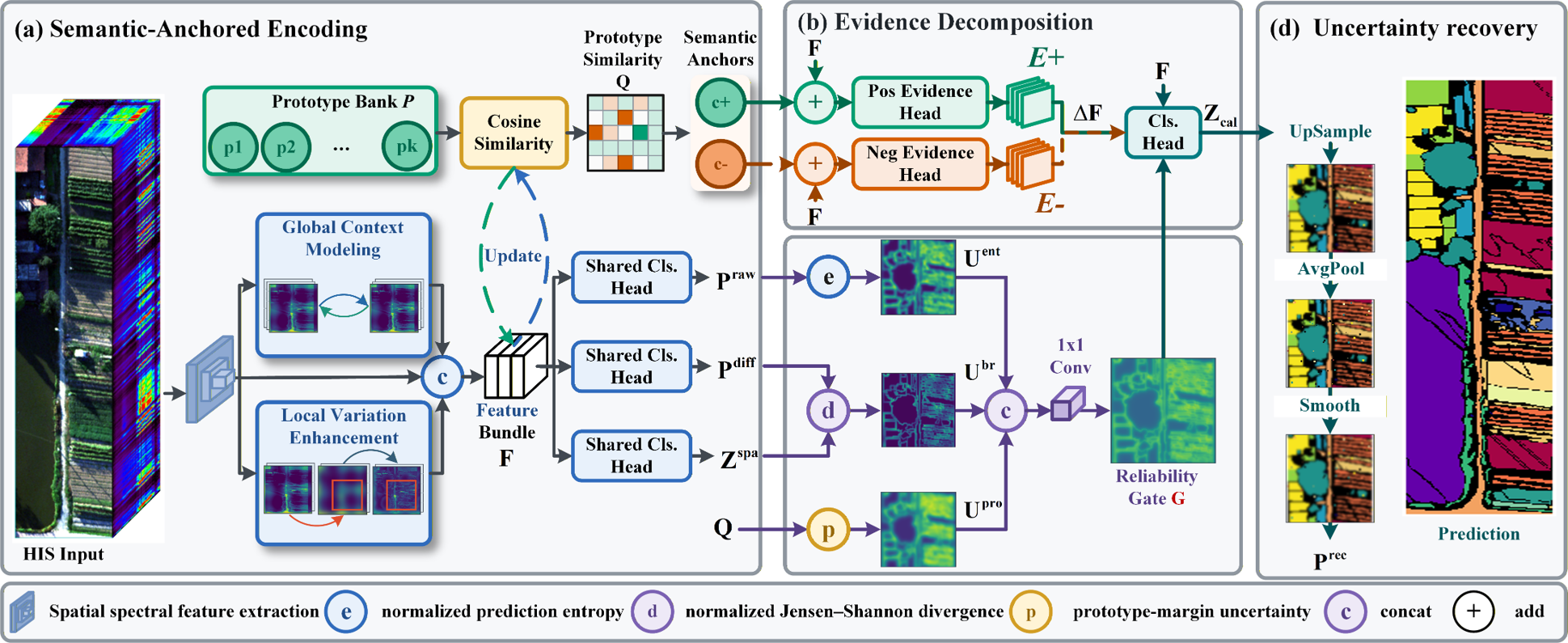}
\caption{Overall architecture of PNEC-Mamba. The framework sequentially performs semantic-anchored encoding, evidence decomposition, reliability gate, uncertainty recovery.}
\label{fig:architecture}
\end{figure*}

\subsection*{Overview}

PNEC-Mamba treats hyperspectral image classification as a pixel-level evidence reliability modeling problem rather than a direct feature-to-label mapping process. Given spectral–spatial representations extracted by a full-image state-space encoder, the framework first establishes adaptive class references through a dynamic prototype bank. Based on pixel–prototype relationships, the model identifies discriminative evidence and conflicting information associated with different classes. A multi-source uncertainty estimation strategy is then employed to evaluate pixel-level reliability and guide selective evidence calibration, where unreliable regions receive stronger adjustment while reliable predictions are preserved. Finally, a spatial consistency refinement step is applied to recover local structures and improve boundary coherence. The overall framework is illustrated in Figure~\ref{fig:architecture}. The overall evidence calibration process is formulated as

\begin{equation}
F_i^{\mathrm{cal}}=F_i+G_i\left(\alpha E_i^{+}-\beta E_i^{-}\right),
\label{eq:overview}
\end{equation}

\noindent where $F_i$ and $F_i^{\mathrm{cal}}$ are the basic and calibrated features of pixel $i$, $E_i^{+}$ and $E_i^{-}$ denote positive and negative evidence, and $G_i$ controls the calibration strength according to pixel reliability.

\subsection*{State-Space Encoding and Prototype Anchoring}

To obtain globally consistent pixel representations for evidence reliability modeling, PNEC-Mamba processes the full hyperspectral image rather than cropped patches. A 1 × 1 convolution is first applied for spectral channel embedding, followed by normalization and nonlinear activation. A subsequent convolution and pooling operation generates a compact feature map, reducing sequence length while maintaining essential spectral–spatial structures for efficient state-space modeling.

The encoder adopts two complementary branches to capture different aspects of spectral–spatial information. The spatial context branch models long-range pixel dependencies through Mamba-based sequence modeling, producing $F^{\mathrm{spa}}$. The differential enhancement branch captures local variations by modeling differences between pixels and their surrounding context, producing $F^{\mathrm{diff}}$. Both branches also produce auxiliary logits $Z^{\mathrm{spa}}$ and $Z^{\mathrm{diff}}$, which are later used to estimate branch disagreement. The two branch features are fused with the residual representation to obtain the initial pixel representation:

\begin{equation}
	F=\omega_{\mathrm{spa}}F^{\mathrm{spa}}+\omega_{\mathrm{diff}}F^{\mathrm{diff}}+F^0,
\end{equation}
where $F^0$ is the residual feature before the Mamba branches, and $\omega_{\mathrm{spa}}$ and $\omega_{\mathrm{diff}}$ are normalized learnable weights. 
The resulting representation is used to generate the initial prediction logits $Z^{\mathrm{raw}}$, which serve as the reference prediction for subsequent reliability estimation and evidence calibration.\citep{li2026diffmamba}.

\subsection{Prototype-Guided Positive-Negative Evidence Construction}

To establish class-level references for evidence analysis, a dynamic class prototype bank $P=\{p_1,p_2,\ldots,p_K\}$ is maintained during training. Here, $K$ is the number of classes, and $p_k \in \mathbb{R}^{D}$ denotes the semantic anchor of class $k$. At each iteration, the normalized class mean $\hat{p}_k$ is computed from labeled pixel representations, and each prototype is updated by momentum to adapt to the evolving feature space:
\begin{equation}
p_k \leftarrow \operatorname{Norm}\left(m p_k + (1-m)\hat{p}_k\right),
\end{equation}
where $m$ is the momentum coefficient, and $\operatorname{Norm}(\cdot)$ denotes $L_2$ normalization. An uninitialized prototype is set to the normalized class mean of the current iteration. Through momentum updates, the prototypes gradually adapt to the evolving feature space and provide dynamic class references during training.

After flattening the spatial dimensions of $F$, let $F_i\in\mathbb{R}^{D}$ denote the feature of pixel $i$, where $i\in\{1,\ldots,N\}$ and $N=hw$. Its scaled cosine similarity to prototype $p_k$ is
\begin{equation}
S_{i,k}=\tau \cdot \frac{F_i^{\top}p_k}{\lVert F_i\rVert\lVert p_k\rVert},
\end{equation}
where $k\in\{1,\ldots,K\}$ is the class index and $\tau>0$ is a learnable similarity scale. Collecting the similarities to all prototypes gives $S_i=[S_{i,1},\ldots,S_{i,K}]^{\top}\in\mathbb{R}^{K}$. The prototype relationship distribution is then obtained as
\begin{equation}
\begin{aligned}
Q_i &= \operatorname{Softmax}(S_i), \\
Q_{i,k} &= \frac{\exp(S_{i,k})}{\sum_{j=1}^{K}\exp(S_{i,j})}.
\end{aligned}
\end{equation}
This distribution provides class-relative information for subsequent evidence construction and prototype-margin uncertainty estimation.


For pixel $i$, the positive anchor is selected as the ground-truth class prototype $p_{y_i}$ for labeled samples during training and the predicted class $p_{\hat{y}_i}$ at inference, where $\hat{y}_i = \operatorname*{argmax}_{k} Q_{i,k}$. The corresponding
positive context is

\begin{equation}
C_i^+=\sum_{k=1}^{K}a_{i,k}^+p_k,
\end{equation}
where $a_i^+$ is a one-hot positive prototype weight. Negative prototypes are selected from the most competitive non-target classes according to the prototype similarity distribution. Their normalized similarities are used to construct the negative prototype context :
\begin{equation}
C_i^-=\sum_{k=1}^{K}a_{i,k}^-p_k.
\end{equation}

Two lightweight evidence heads are introduced to learn positive and negative evidence representations, respectively.
\begin{align}
E_i^+ &= g_+([F_i,F_i^{\mathrm{spa}},F_i^{\mathrm{diff}},C_i^+]),\\
E_i^- &= g_-([F_i,F_i^{\mathrm{spa}},F_i^{\mathrm{diff}},C_i^-]),
\end{align}
where $[\cdot]$ denotes channel concatenation. The final layers of both heads are zero-initialized so that evidence calibration is gradually introduced without disrupting the initial classification behavior. The resulting evidence adjustment is defined as:
\begin{equation}
\Delta F_i=\alpha E_i^+-\beta E_i^-,
\end{equation}
where $\alpha$ and $\beta$ are non-negative scaling coefficients. Positive evidence enhances class-consistent responses, while negative evidence suppresses responses associated with confusing classes.

\subsection*{Multi-Source Uncertainty Estimation and Evidence Calibration}

To selectively calibrate unreliable pixels while preserving reliable predictions, we estimate pixel-level uncertainty from three complementary sources. First, normalized predictive entropy is

\begin{equation}
U_i^{\mathrm{ent}}=-\frac{1}{\log K}\sum_{k=1}^{K}P_{i,k}^{\mathrm{raw}}\log P_{i,k}^{\mathrm{raw}},
\end{equation}
where $P_i^{\mathrm{raw}}=\operatorname{Softmax}(Z_i^{\mathrm{raw}})$. Second,representation inconsistency between the two complementary branches is measured using normalized Jensen–Shannon divergence: $\bar P_i=(P_i^{\mathrm{spa}}+P_i^{\mathrm{diff}})/2$. Then
\begin{equation}
U_i^{\mathrm{br}}=\frac{\operatorname{KL}(P_i^{\mathrm{spa}}\Vert\bar P_i)+\operatorname{KL}(P_i^{\mathrm{diff}}\Vert\bar P_i)}{2\log2}.
\end{equation}
Third, prototype ambiguity is measured by the margin between the highest and second-highest prototype probabilities.

\begin{equation}
U_i^{\mathrm{pro}}=1-\left(q_i^{(1)}-q_i^{(2)}\right).
\end{equation}

The three maps are concatenated and fused by a $1\times1$ convolution to obtain the reliability guidance map $\widetilde G_i$. A gate floor ensures sufficient calibration while preserving relative uncertainty:
\begin{equation}
G_i=\rho+(1-\rho)\widetilde G_i,
\end{equation}
where $\rho\in[0,1]$. The final evidence-calibrated representation is obtained as
\begin{equation}
F_i^{\mathrm{cal}}=F_i+G_i(\alpha E_i^+-\beta E_i^-).
\end{equation}
The same classification head maps $F^{\mathrm{cal}}$ to calibrated logits $Z^{\mathrm{cal}}$.

\subsection*{Spatial Consistency Refinement}

Although evidence calibration improves unreliable predictions, the resolution reduction introduced during state-space encoding may affect fine-grained spatial structures. Direct upsampling may introduce isolated predictions and fragmented boundaries, especially in heterogeneous regions. We therefore refine predictions in the full-resolution probability space using local consistency. Let
\begin{equation}
P^{\mathrm{up}}=\operatorname{Softmax}(\operatorname{Up}(Z^{\mathrm{cal}})),
\end{equation}
where $\operatorname{Up}(\cdot)$ denotes bilinear upsampling. Window averaging gives
\begin{equation}
\bar P^{\mathrm{up}}=\operatorname{AvgPool}_k(P^{\mathrm{up}}),
\end{equation}
and the recovered probability is
\begin{equation}
P^{\mathrm{rec}}=(1-\eta)P^{\mathrm{up}}+\eta\bar P^{\mathrm{up}},
\end{equation}
where $\eta$ controls recovery strength. The final label is obtained from the maximum component of $ P^{\mathrm{rec}}$.

\subsection*{Training Objective}

PNEC-Mamba is optimized with classification supervision, auxiliary representation constraints, and a reliability-aware consistency regularizer. Let
$\Omega_L$ denote the set of labeled training pixels and $N_L=|\Omega_L|$. The calibrated logits provide the primary classification supervision. Meanwhile the two encoder branches and the prototype similarities are supervised through auxiliary objectives:
\begin{align}
	\mathcal{L}_{\mathrm{ce}}
	&=
	\frac{1}{N_L}
	\sum_{i\in\Omega_L}
	\operatorname{CE}(Z_i^{\mathrm{cal}},y_i),\\
	\mathcal{L}_{\mathrm{br}}
	&=
	\frac{1}{N_L}
	\sum_{i\in\Omega_L}
	\left[
	\operatorname{CE}(Z_i^{\mathrm{spa}},y_i)
	+
	\operatorname{CE}(Z_i^{\mathrm{diff}},y_i)
	\right],\\
	\mathcal{L}_{\mathrm{pro}}
	&=
	\frac{1}{N_L}
	\sum_{i\in\Omega_L}
	\operatorname{CE}(S_i,y_i),
\end{align}
where $S_i=[S_{i,1},\ldots,S_{i,K}]$ denotes the
prototype-similarity logits of pixel $i$.

To ensure that evidence calibration focuses on unreliable regions without disturbing reliable predictions, we introduce a reliability-aware consistency loss. Let $P_i^{\mathrm{raw}}=\operatorname{Softmax}(Z_i^{\mathrm{raw}})$ and $P_i^{\mathrm{cal}}=\operatorname{Softmax}(Z_i^{\mathrm{cal}})$.
The consistency objective is
\begin{equation}
	\mathcal{L}_{\mathrm{con}}
	=
	\frac{1}{N_C}
	\sum_{i\in\Omega_C}
	(1-U_i^{\mathrm{fuse}})
	\operatorname{KL}
	\left(
	P_i^{\mathrm{raw}}
	\Vert
	P_i^{\mathrm{cal}}
	\right),
\end{equation}
where $\widetilde G_i\in[0,1]$ is the fused uncertainty score,
$\Omega_C$ is the set of pixels used for consistency
regularization, and $N_C=|\Omega_C|$. Consequently, reliable
pixels with low uncertainty receive stronger consistency
constraints, whereas uncertain pixels retain greater freedom
for evidence calibration.

The complete training objective is
\begin{equation}
	\mathcal{L}
	=
	\mathcal{L}_{\mathrm{ce}}
	+
	\lambda_{\mathrm{br}}\mathcal{L}_{\mathrm{br}}
	+
	\lambda_{\mathrm{pro}}\mathcal{L}_{\mathrm{pro}}
	+
	\lambda_{\mathrm{con}}\mathcal{L}_{\mathrm{con}},
\end{equation}
where $\lambda_{\mathrm{br}}$, $\lambda_{\mathrm{pro}}$, and
$\lambda_{\mathrm{con}}$ control the contributions of the
corresponding objectives.

\section{Experiments}

\subsection*{Experimental Setup}

Three benchmark hyperspectral datasets are used to evaluate PNEC-Mamba: Pavia University (UP), Houston 2013 \citep{debes2014houston}, and WHU-Hi-HanChuan \citep{zhong2020whuhi}. These datasets cover two urban scenes and one agricultural scene and differ substantially in spatial resolution, spectral dimensionality, scene extent, and land-cover composition, as summarized in Table~\ref{tab:datasets}. Such diversity provides comprehensive evaluations under different levels of spectral ambiguity, class similarity, and spatial heterogeneity. These properties provide diverse conditions for evaluating classification performance in complex boundaries and heterogeneous regions.

\begin{table}[t]
\centering
\tablefont
\setlength{\tabcolsep}{1.5mm}
\begin{tabular}{lcccc}
\toprule
Dataset & GSD & Bands & Classes & Train/Class \\
\midrule
UP & 1.3 m & 103 & 9 & 15 \\
Houston & 2.5 m & 144 & 15 & 15 \\
HanChuan & 0.109 m & 274 & 16 & 30 \\
\bottomrule
\end{tabular}
\caption{Summary of datasets and training samples.}
\label{tab:datasets}
\end{table}

All experiments were implemented with PyTorch 1.13.1 and CUDA 11.7 on an NVIDIA A800 GPU with 80 GB of memory. Fixed training and test partitions were used, and each experiment was repeated ten times. Mean values and standard deviations are reported. Classification performance is evaluated using overall accuracy (OA), average accuracy (AA), the Kappa coefficient, and class-wise accuracy. Additional ablation and mechanism analyses are conducted to investigate the contribution of each component and the effectiveness of evidence reliability modeling. Detailed training configurations are provided in the Appendix.

\subsection*{Comparative Experiments}
PNEC-Mamba is compared with representative CNN-, Transformer-, and Mamba-based hyperspectral classification methods. The selected baselines cover convolutional models (DBMA and DBDA) \citep{li2019dbma,li2020dbda}, Transformer-based models (GAHT and MASSFormer) \citep{mei2022gaht,sun2024massformer}, and recent state-space models (S2Mamba, MambaLG, MambaHSI, and MambaMoE) \citep{wang2025s2mamba,pan2025mambalg,li2024mambahsi,xu2026mambamoe}. These methods provide comprehensive comparisons across different representation learning paradigms.

\begin{table}[t]
\centering
\tablefont
\setlength{\tabcolsep}{1.15pt}
\renewcommand{\arraystretch}{0.92}
\begin{tabular}{@{}lccc||ccc@{}}
\toprule
& \multicolumn{3}{c||}{Pavia University (UP)} & \multicolumn{3}{c}{Houston} \\
\cmidrule(lr){2-4}\cmidrule(l){5-7}
Method & OA & AA & $\kappa$ & OA & AA & $\kappa$ \\
\midrule
DBMA & \res{92.10}{3.49} & \res{92.96}{2.86} & \res{89.72}{4.47} & \res{85.01}{5.08} & \res{87.02}{4.95} & \res{83.80}{5.50} \\
DBDA & \res{94.19}{1.32} & \res{95.17}{1.08} & \res{92.41}{1.71} & \res{89.25}{1.38} & \res{90.74}{1.15} & \res{88.38}{1.50} \\
GAHT & \res{85.36}{2.81} & \res{86.84}{3.93} & \res{81.18}{3.41} & \res{87.65}{0.68} & \res{89.88}{0.54} & \res{86.66}{0.74} \\
MASSFormer & \res{94.03}{0.73} & \res{94.03}{0.86} & \res{92.17}{0.95} & \res{91.23}{0.65} & \res{92.75}{0.62} & \res{90.53}{0.70} \\
S2Mamba & \res{93.14}{0.87} & \res{94.23}{0.59} & \res{91.06}{1.11} & \res{82.81}{2.08} & \res{84.87}{1.52} & \res{81.44}{2.23} \\
MambaLG & \res{93.59}{0.87} & \res{93.66}{0.56} & \res{91.59}{1.11} & \res{89.86}{0.99} & \res{91.49}{0.82} & \res{89.04}{1.07} \\
MambaHSI & \res{92.82}{0.65} & \res{93.18}{0.48} & \res{90.60}{0.83} & \res{89.32}{0.51} & \res{90.92}{0.43} & \res{88.46}{0.55} \\
MambaMoE & \res{94.71}{1.41} & \bestres{96.73}{0.76} & \res{93.09}{1.78} & \res{89.82}{1.07} & \res{91.51}{0.83} & \res{89.01}{1.15} \\
\rowcolor{pneccol}
PNEC-Mamba & \bestres{96.77}{0.43} & \res{95.84}{0.55} & \bestres{95.73}{0.56} & \bestres{92.18}{0.38} & \bestres{93.49}{0.31} & \bestres{91.55}{0.41} \\
\bottomrule
\end{tabular}
\caption{Aggregate classification results on Pavia University and Houston.}
\label{tab:aggregate_main}
\end{table}

\begin{table*}[t]
	\centering
	\tablefont
	\setlength{\tabcolsep}{2.8pt}
	\renewcommand{\arraystretch}{0.90}
	
	\newcommand{\hcres}[2]{%
		\shortstack{$#1$\\$\pm #2$}%
	}
	\newcommand{\hcbestres}[2]{%
		\shortstack{$\mathbf{#1}$\\$\mathbf{\pm #2}$}%
	}
	
	\begin{tabular}{@{}c||cc||cc||cccc||>{\columncolor{pneccol}}c@{}}
		\toprule
		& \multicolumn{2}{c||}{CNN-based}
		& \multicolumn{2}{c||}{Transformer-based}
		& \multicolumn{4}{c||}{Mamba-based}
		& \cellcolor{pneccol}{Proposed} \\
		
		\cmidrule(lr){2-3}
		\cmidrule(lr){4-5}
		\cmidrule(lr){6-9}
		\cmidrule(l){10-10}
		
		Class
		& DBMA
		& DBDA
		& GAHT
		& MASSFormer
		& S2Mamba
		& MambaLG
		& MambaHSI
		& MambaMoE
		& PNEC-Mamba \\
		
		\midrule
		
		1  & 90.75 & 93.68 & 87.58 & 92.92 & 92.35 & 88.54 & 92.86 & 94.39 & \textbf{98.69} \\
		2  & 85.31 & 87.38 & 83.51 & 87.73 & 87.90 & 84.24 & 78.08 & 86.64 & \textbf{89.63} \\
		3  & 92.88 & 95.70 & 88.48 & 93.72 & 95.02 & 94.01 & 98.03 & 92.02 & \textbf{99.43} \\
		4  & 96.79 & 97.61 & 96.39 & 98.90 & 97.31 & 96.86 & 98.94 & 94.03 & \textbf{99.84} \\
		5  & \textbf{99.85} & 99.74 & 98.32 & 99.60 & 99.55 & 99.85 & 99.70 & 99.32 & 99.68 \\
		6  & 82.53 & \textbf{86.94} & 73.46 & 83.40 & 83.48 & 80.30 & 62.55 & 61.75 & 32.11 \\
		7  & 95.11 & 96.65 & 89.98 & 94.67 & 93.86 & 92.18 & 92.10 & 78.50 & \textbf{97.66} \\
		8  & 74.27 & 73.67 & 71.34 & 75.27 & \textbf{79.94} & 77.87 & 56.97 & 78.93 & 77.49 \\
		9  & 92.45 & \textbf{96.51} & 87.85 & 90.73 & 92.35 & 89.74 & 91.67 & 87.34 & 92.65 \\
		10 & 96.17 & 96.62 & 97.32 & 96.95 & \textbf{98.06} & 95.99 & 92.88 & 93.67 & 94.69 \\
		11 & 94.43 & 96.47 & 90.53 & 96.43 & 96.86 & 94.36 & 91.47 & 93.49 & \textbf{97.94} \\
		12 & 92.48 & 88.98 & 85.23 & 90.54 & 84.79 & \textbf{92.66} & 82.35 & 85.24 & 84.59 \\
		13 & 81.98 & \textbf{88.11} & 83.85 & 80.61 & 83.11 & 79.76 & 84.43 & 82.98 & 86.16 \\
		14 & 82.76 & 86.62 & 82.17 & 86.57 & 83.39 & 87.46 & 90.35 & 89.32 & \textbf{95.60} \\
		15 & 99.60 & \textbf{99.93} & 99.10 & 97.68 & 99.87 & 98.54 & 97.91 & 99.30 & 96.34 \\
		16 & 98.14 & 98.28 & 98.46 & 98.55 & 97.68 & 97.68 & 98.13 & 93.16 & \textbf{99.17} \\
		
		\midrule
		
		OA
		& \hcres{91.20}{0.87}
		& \hcres{92.87}{1.00}
		& \hcres{89.45}{0.89}
		& \hcres{92.35}{0.37}
		& \hcres{92.26}{0.27}
		& \hcres{90.92}{1.47}
		& \hcres{89.67}{1.14}
		& \hcres{90.03}{3.61}
		& \hcbestres{94.11}{0.43} \\
		
		AA
		& \hcres{90.97}{0.77}
		& \hcbestres{92.68}{1.33}
		& \hcres{88.35}{1.02}
		& \hcres{91.52}{0.48}
		& \hcres{91.60}{0.23}
		& \hcres{90.63}{1.03}
		& \hcres{88.03}{0.98}
		& \hcres{88.13}{3.89}
		& \hcres{90.10}{1.16} \\
		
		$\kappa$
		& \hcres{89.74}{1.00}
		& \hcres{91.69}{1.16}
		& \hcres{87.71}{1.04}
		& \hcres{91.07}{0.43}
		& \hcres{90.97}{0.31}
		& \hcres{89.43}{1.69}
		& \hcres{87.93}{1.33}
		& \hcres{88.42}{4.11}
		& \hcbestres{93.10}{0.50} \\
		
		\bottomrule
	\end{tabular}
	
	\caption{Classification results on WHU-Hi-HanChuan.
		Class-wise entries are mean values; OA, AA, and $\kappa$
		are reported as mean $\pm$ standard deviation.}
	\label{tab:hanchuan_main}
	
\end{table*}

The quantitative results in Tables~\ref{tab:aggregate_main} and~\ref{tab:hanchuan_main} demonstrate that PNEC-Mamba achieves the best OA and $\kappa$ on all three datasets. On UP, PNEC-Mamba improves OA and $\kappa$ over the strongest baseline by 2.06 and 2.64 percentage points, respectively. On Houston, it outperforms MASSFormer by 0.95, 0.74, and 1.02 percentage points in OA, AA, and $\kappa$, respectively. On WHU-Hi-HanChuan, the improvements over the strongest baseline are 1.24 and 1.41 percentage points in OA and $\kappa$, respectively. These consistent gains demonstrate the robustness of PNEC-Mamba across urban and agricultural scenes with different spatial resolutions and land-cover structures.

\begin{figure*}[t]
\centering
\includegraphics[width=0.56\textwidth]{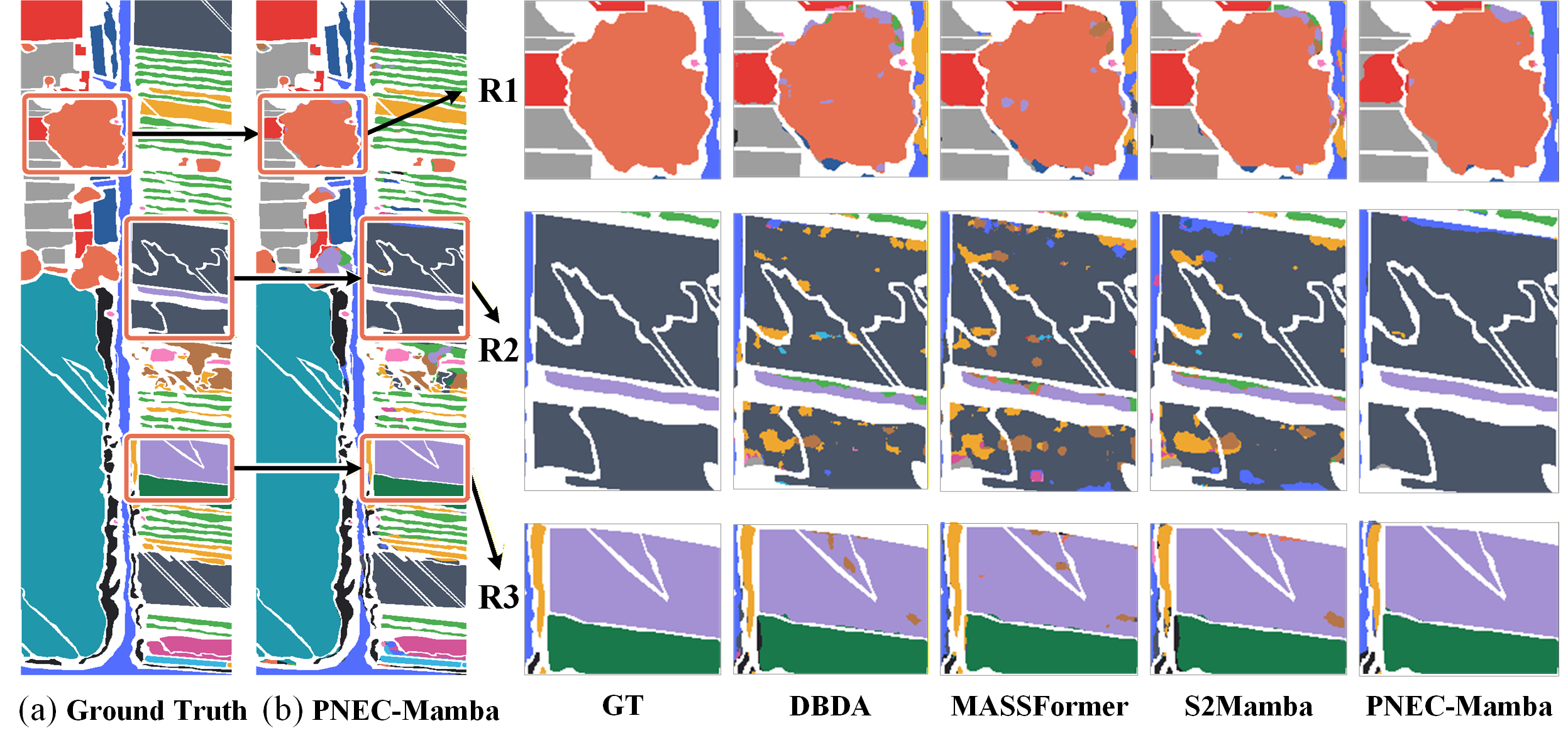}
\caption{Classification maps on WHU-Hi-HanChuan: (a) ground truth, (b) PNEC-Mamba.}
\label{fig:hanchuan_maps}
\end{figure*}

Figure~\ref{fig:hanchuan_maps} further shows that PNEC-Mamba produces more spatially coherent predictions, with fewer isolated errors and cleaner boundaries, especially in heterogeneous regions where conflicting evidence is more prevalent. The corresponding classification maps for UP and Houston are provided in the appendix. These quantitative and qualitative results support the effectiveness of separating supportive evidence from competitive interference for robust hyperspectral image classification.

\subsection*{Computational Efficiency}

We further evaluate computational efficiency using model parameters, training time, and full-scene inference time. Training time is averaged over ten aligned runs. Full-scene inference predicts every pixel in the image and includes patch extraction for patch-based baselines. Table~\ref{tab:efficiency} summarizes the comparison.

\begin{table*}[t]
\centering
\tablefont
\setlength{\tabcolsep}{2.0pt}
\renewcommand{\arraystretch}{0.90}
\begin{tabular}{@{}lccc||ccc||ccc@{}}
\toprule
Method & \multicolumn{3}{c||}{UP} & \multicolumn{3}{c||}{Houston} & \multicolumn{3}{c}{WHU-Hi-HanChuan} \\
 & \shortstack{Params\\(M)} & \shortstack{Training\\Time (s)} & \shortstack{Full\\Inference (s)} & \shortstack{Params\\(M)} & \shortstack{Training\\Time (s)} & \shortstack{Full\\Inference (s)} & \shortstack{Params\\(M)} & \shortstack{Training\\Time (s)} & \shortstack{Full\\Inference (s)} \\
\midrule
DBMA & 0.321 & 64.89$\pm$4.17 & 29.1694 & 0.443 & 100.05$\pm$1.76 & 120.6094 & 0.821 & 540.59$\pm$30.85 & 87.3174 \\
DBDA & 0.203 & 66.27$\pm$5.03 & 22.8380 & 0.280 & 99.01$\pm$2.56 & 110.8560 & 0.517 & 468.45$\pm$27.00 & 94.7410 \\
GAHT & 0.927 & 15.32$\pm$1.04 & 17.6822 & 0.974 & 18.64$\pm$1.46 & 53.8699 & 1.123 & 81.56$\pm$6.17 & 41.8760 \\
MASSFormer & 0.314 & 23.26$\pm$2.08 & 13.3389 & 0.315 & 36.61$\pm$2.00 & 38.5535 & 0.315 & 123.03$\pm$4.47 & 20.6132 \\
S2Mamba & 0.120 & 15.56$\pm$0.31 & 13.4862 & 0.119 & 20.59$\pm$0.39 & 42.6287 & 0.132 & 96.65$\pm$3.54 & 41.0010 \\
MambaLG & 0.187 & 69.55$\pm$0.37 & 0.1164 & 0.297 & 260.71$\pm$0.42 & 0.4264 & 0.198 & 207.07$\pm$0.30 & 0.2125 \\
MambaHSI & 0.412 & 248.04$\pm$51.34 & 0.1689 & 0.418 & 584.75$\pm$55.84 & 0.5522 & 0.435 & 512.88$\pm$0.35 & 0.3062 \\
MambaMoE & 0.687 & 271.87$\pm$47.58 & 0.0801 & 0.695 & 440.37$\pm$36.98 & 0.2506 & 0.712 & 482.59$\pm$50.76 & 0.1610 \\
\rowcolor{pneccol}
PNEC-Mamba & 0.646 & 35.73$\pm$1.90 & 0.0149 & 0.654 & 82.13$\pm$4.67 & 0.0443 & 2.432 & 280.97$\pm$69.45 & 0.0508 \\
\bottomrule
\end{tabular}
\caption{Computational efficiency comparison on the three benchmark datasets.}
\label{tab:efficiency}
\end{table*}

PNEC-Mamba is not the smallest model in terms of parameters. However, it provides a favorable balance between accuracy and efficiency. Its full-scene inference is consistently the fastest among all compared methods. This advantage is especially clear against patch-based CNN and Transformer baselines, which require repeated local-window evaluation. Compared with full-image Mamba baselines, PNEC-Mamba also keeps the training cost within a practical range while achieving higher OA and $\kappa$.

\subsection*{Ablation Study}

To evaluate the contribution of each component, we conduct controlled ablation experiments under the same experimental settings. The Base model denotes the state-space representation model without evidence reliability modeling. The w/o EC variant removes prototype-guided positive–negative evidence construction, while the w/o RC variant removes reliability-aware calibration. The complete PNEC-Mamba integrates all components.

\begin{table}[t]
\centering
\tablefont
\setlength{\tabcolsep}{2.1pt}
\renewcommand{\arraystretch}{0.94}
\begin{tabular}{@{}llccc>{\columncolor{pneccol}}c@{}}
\toprule
Dataset & Metric & Base & w/o EC &  w/o RC & \cellcolor{pneccol}{PNEC-Mamba} \\
\midrule
UP
& OA       & 95.63 & 97.00 & 95.83 & \gain{97.39}{1.76} \\
& AA       & 96.07 & 96.43 & 96.35 & \gain{96.72}{0.65} \\
& $\kappa$ & 94.25 & 96.03 & 94.51 & \gain{96.55}{2.30} \\
\midrule
Houston
& OA       & 92.07 & 92.08 & 92.30 & \gain{92.47}{0.40} \\
& AA       & 93.39 & 93.33 & \textbf{93.60} & \gain{93.60}{0.21} \\
& $\kappa$ & 91.43 & 91.43 & 91.68 & \gain{91.86}{0.43} \\
\midrule
HanChuan
& OA       & 92.73 & 94.65 & 93.10 & \gain{94.79}{2.06} \\
& AA       & 88.67 & 89.30 & 88.82 & \gain{90.68}{2.01} \\
& $\kappa$ & 91.49 & 93.73 & 91.93 & \gain{93.90}{2.41} \\
\bottomrule
\end{tabular}
\caption{Ablation results.}
\label{tab:ablation}
\end{table}

As shown in Table~\ref{tab:ablation}, PNEC achieves the best performance across all datasets and metrics. The consistent gains demonstrate that the proposed evidence reliability modeling provides substantial discrimination beyond the basic spectral--spatial representation, with particularly pronounced improvements on UP and HanChuan. Removing RC causes clear performance drops, indicating that reliability estimation is essential for selectively calibrating uncertain regions while avoiding unnecessary modification of reliable predictions. Removing EC consistently decreases performance, demonstrating that explicitly separating positive and negative evidence improves discrimination in ambiguous pixels. Overall, EC and RC provide complementary benefits, and their integration yields the strongest and most consistent performance.

\subsection*{Evidence Reliability Analysis and Findings}

We further investigate whether the proposed reliability estimation can effectively identify pixels requiring evidence calibration, and the following empirical findings were observed. 

\textbf{Finding 1: Incorrectly  pixels exhibit higher uncertainty.} According to the results in Table~\ref{tab:uncertainty}, incorrectly classified pixels exhibit higher uncertainty than correctly classified pixels, indicating that the estimated reliability effectively distinguishes stable and unstable predictions. This is further supported by the fact that the most uncertain 20\% of pixels cover 73.29\%--82.98\% of all errors. Moreover, the error AUROC values remain high, ranging from 0.8891 to 0.9157, demonstrating the capability of uncertainty estimation to localize unreliable decisions.

\begin{figure}[t]
\centering
\includegraphics[width=0.8\columnwidth]{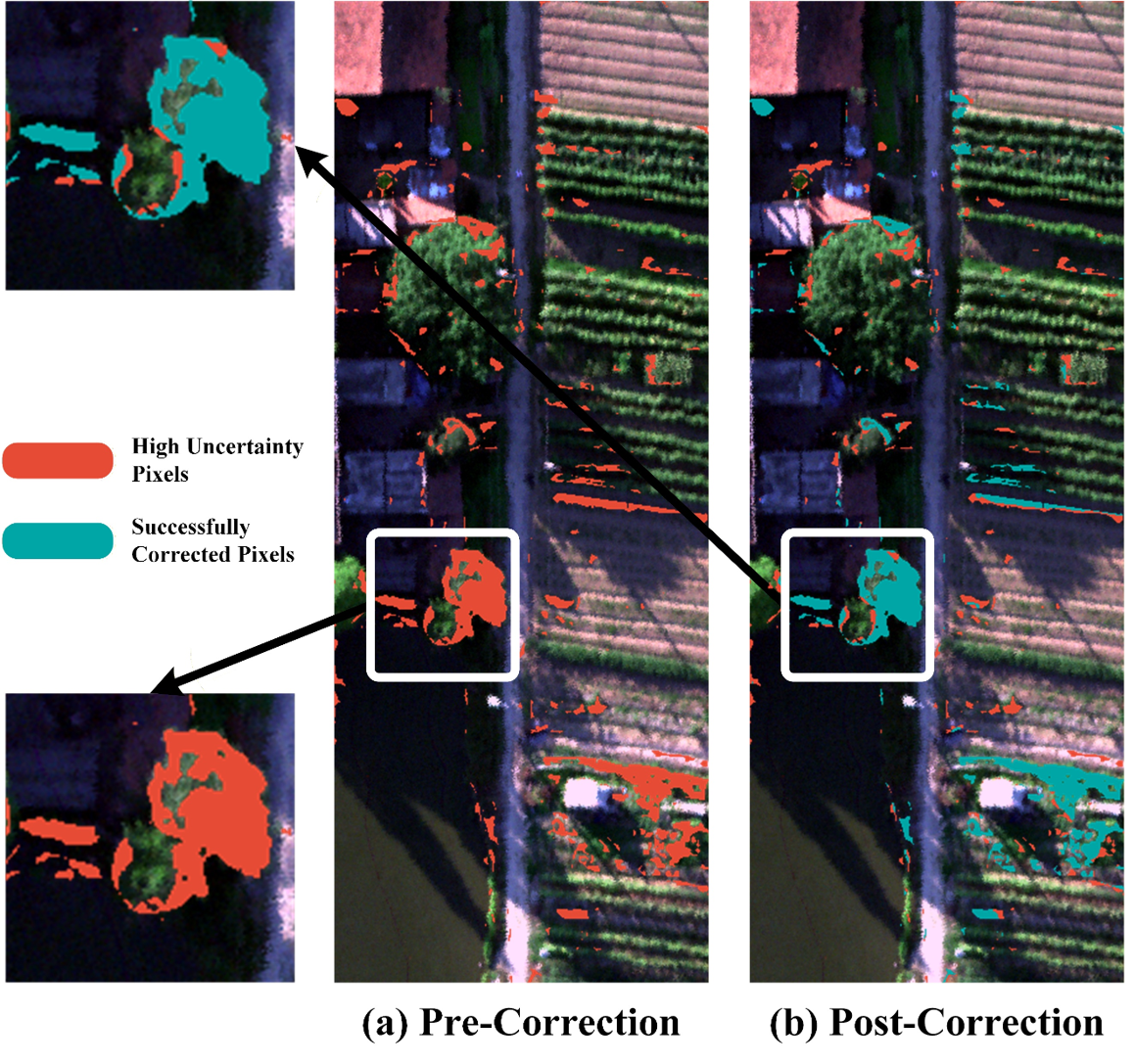}
\caption{Spatial visualization of high-uncertainty errors and successful corrections on WHU-Hi-HanChuan.}
\label{fig:spatial_reliability}
\end{figure}

\textbf{Finding 2: Unreliable evidence is not uniformly distributed.} Figure~\ref{fig:spatial_reliability} visualizes reliability-guided correction on WHU-Hi-HanChuan. Orange marks misclassified pixels within the top 20\% uncertainty region, while cyan marks those successfully corrected after evidence calibration and local consistency recovery. High-uncertainty errors are concentrated around field boundaries, heterogeneous vegetation, narrow structures, and mixed regions. Many of these high-risk pixels are correctly recovered after refinement, particularly in the enlarged regions. This spatial correspondence demonstrates that the estimated reliability successfully identifies ambiguous regions where evidence calibration is most beneficial.

\textbf{Finding 3: The factor limiting model accuracy is no longer whether the model can extract more features, but the purity of the features.} Comparative experiments and ablation studies show that, under the same training samples, the best models in the CNN, Transformer, and Mamba frameworks achieve accuracies that differ by less than 1\% (92.87\%/92.26\%/92.73\%). This indicates that, regardless of the type or number of features extracted, they are already sufficient for classification. At present, the reason limiting classification accuracy may no longer be feature quantity, but feature quality. Therefore, it is worth further study to distinguish useful features from noise, and to better handle objects with high uncertainty.

\begin{table}[t]
\centering
\tablefont
\setlength{\tabcolsep}{0.6pt}
\renewcommand{\arraystretch}{0.94}
\begin{tabular}{@{}lcccc@{}}
\toprule
Dataset & \shortstack{Correct\\Uncertainty} & \shortstack{Wrong\\Uncertainty} & \shortstack{Error\\AUROC} & \shortstack{Top-20\%\\Error Recall} \\
\midrule
UP & 0.1668 & 0.2852 & 0.8938 & 79.78\% \\
Houston & 0.1705 & 0.3037 & 0.8891 & 73.29\% \\
WHU-Hi-HanChuan & 0.1991 & 0.3869 & 0.9157 & 82.98\% \\
\bottomrule
\end{tabular}
\caption{Reliability-based error localization.}
\label{tab:uncertainty}
\end{table}

\FloatBarrier
\section{Conclusion}

We introduced PNEC-Mamba, a pixel-level evidence reliability modeling framework with prototype-guided positive–negative evidence calibration for hyperspectral image classification. Rather than directly classifying fused representations, PNEC-Mamba explicitly models whether pixel-level evidence contributes to or conflicts with classification decisions. Dynamic class prototypes provide adaptive class references, while prototype competition derives positive and negative evidence from pixel–class relationships. Multi-source reliability estimation then controls selective evidence calibration according to pixel-level reliability. The calibrated outputs are further refined through spatial consistency refinement to improve prediction coherence.

On Pavia University, Houston, and WHU-Hi-HanChuan, PNEC-Mamba achieved OA values of 96.77\%, 92.18\%, and 94.11\%, respectively, and obtained the best OA and $\kappa$ among all compared methods. Ablation studies confirmed the complementary contributions of evidence construction and reliability-aware calibration. Analysis and Findings section presents and supports the scientific findings of this study and suggests promising directions for future research. Future work will focus on cross-scene generalization and more efficient evidence reliability modeling.

\clearpage
\appendix
\section{Appendix}
\subsection{Experimental Details}
The experiments used PyTorch 1.13.1 and CUDA 11.7 on an NVIDIA A800 GPU with 80~GB memory. We used Adam with an initial learning rate of $3\times10^{-4}$. Evidence calibration was disabled for the first 10 epochs, while the prototype bank and auxiliary heads remained active. The loss weights were $\lambda_{br}=0.1$, $\lambda_{pro}=0.1$, and $\lambda_{con}=0.02$. The hidden dimension was 128 for UP and Houston and 256 for WHU-Hi-HanChuan. For these three datasets, the gate floors were 0.5, 0.3, and 0.5; the local recovery windows were 9, 11, and 13; and the recovery strengths were 1.0, 0.85, and 1.0, respectively.

All methods used the same fixed data partitions. With the split seed set to 12345, we randomly selected 15 training pixels per class for UP and Houston and 30 per class for WHU-Hi-HanChuan. All remaining labeled pixels formed the test set. The resulting indices were saved and reused by every method.

Each method was run independently ten times using the following seed sequence:
\[
\begin{gathered}
	\{202501, 202502, 202503, 202504, 202505,\\
	202506, 202507, 202508, 202509, 2025010\}.
\end{gathered}
\]
Before each run, the corresponding seed initialized the random number generators of Python, NumPy, PyTorch, and CUDA. This controlled parameter initialization, data ordering, and stochastic augmentation. Each method retained its architecture-specific optimization settings. We report the mean and standard deviation over the ten runs.

\subsection{Training and Inference Procedure}

Algorithm~\ref{alg:pnec} summarizes the supervised training procedure and the parameter-free full-resolution recovery used during inference.

During the first $T_w=10$ epochs, $s=0$ disables evidence correction. The prototype bank and auxiliary heads remain active to stabilize the feature space and semantic anchors. After warm-up, the calibrated logits $Z^{\mathrm{cal}}$ are supervised on labeled pixels together with the branch, prototype, and consistency objectives.

The recovery probability $P^{\mathrm{rec}}$ is computed only during inference. This parameter-free operation is excluded from back-propagation and improves local coherence at the original image resolution.

\subsection{Detailed Comparison Results}
\label{app:detailed_results}

\begin{figure*}[!t]
	\centering
	\includegraphics[width=\textwidth]{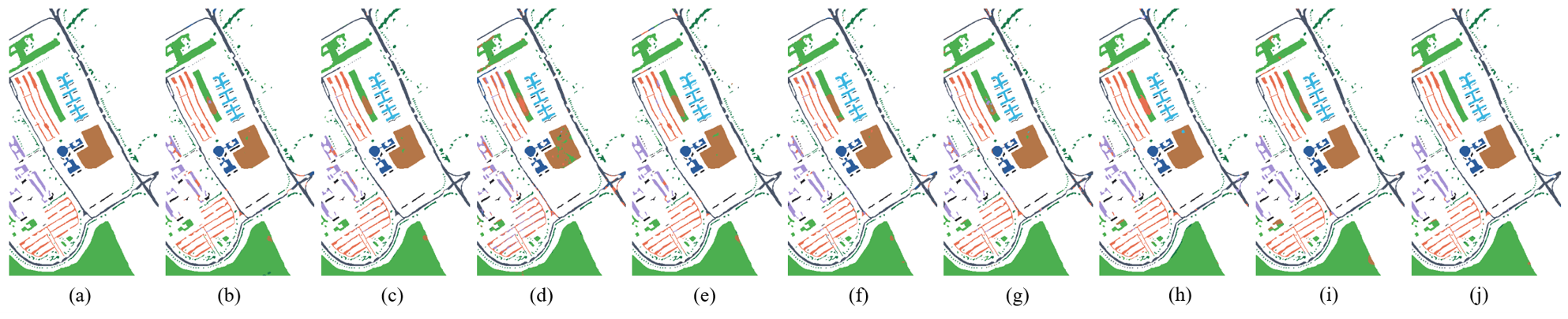}
	\caption{Classification maps on Pavia University: (a) ground truth, (b) DBMA, (c) DBDA, (d) GAHT, (e) MASSFormer, (f) S2Mamba, (g) MambaLG, (h) MambaHSI, (i) MambaMoE, and (j) PNEC-Mamba.}
	\label{fig:up_maps_appendix}
\end{figure*}

\begin{figure*}[!t]
	\centering
	\includegraphics[width=\textwidth]{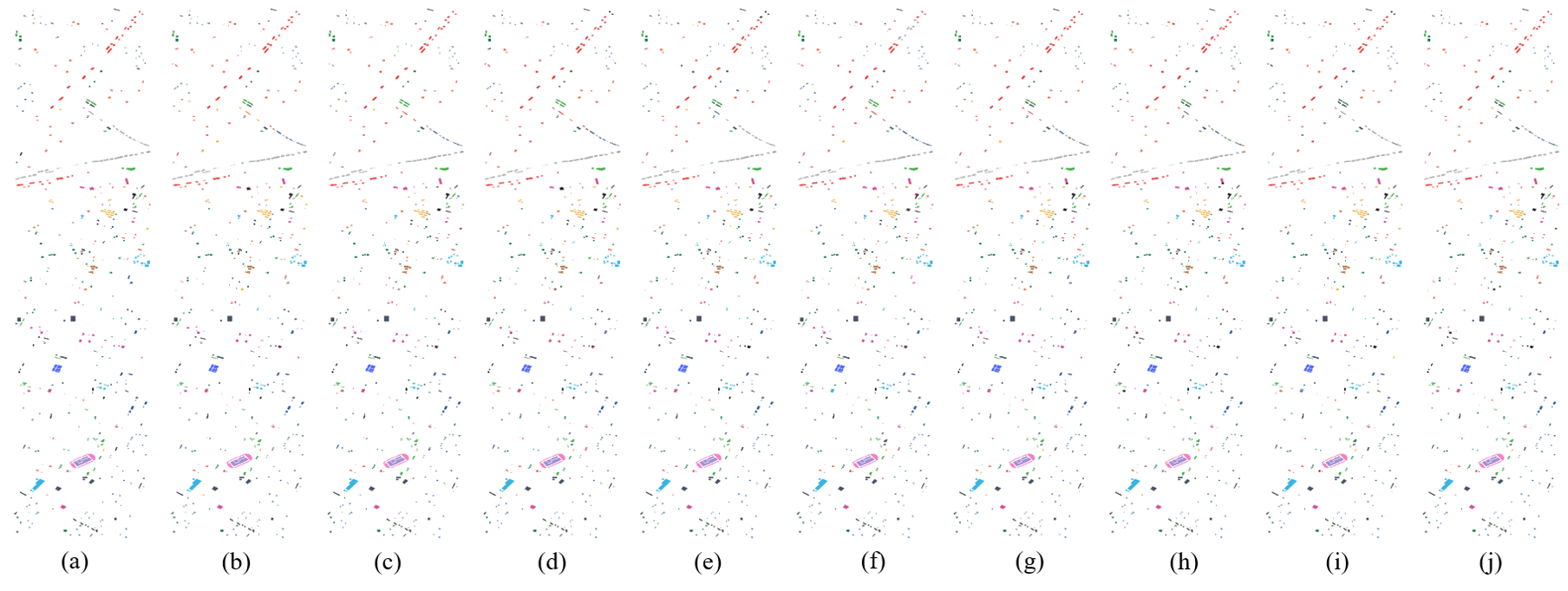}
	\caption{Classification maps on Houston: (a) ground truth, (b) DBMA, (c) DBDA, (d) GAHT, (e) MASSFormer, (f) S2Mamba, (g) MambaLG, (h) MambaHSI, (i) MambaMoE, and (j) PNEC-Mamba.}
	\label{fig:houston_maps_appendix}
\end{figure*}

\begin{figure*}[!t]
	\centering
	\includegraphics[width=\textwidth]{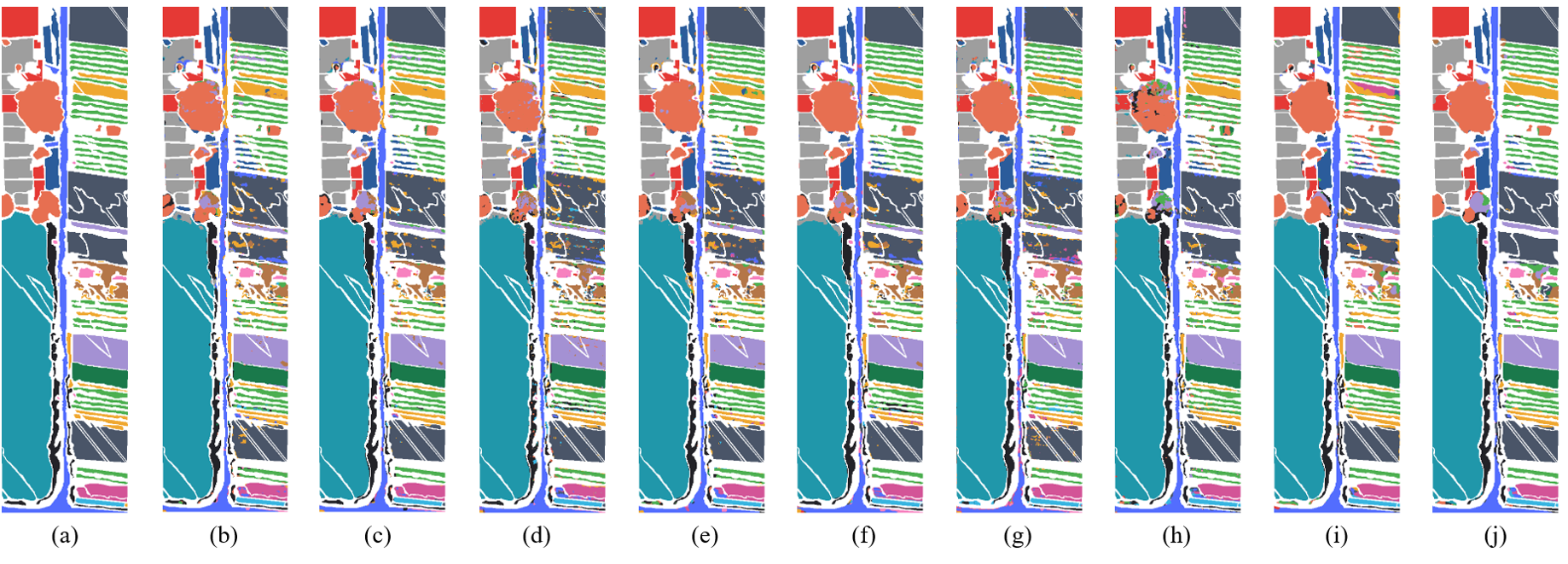}
	\caption{Classification maps on WHU-Hi-HanChuan: (a) ground truth, (b) DBMA, (c) DBDA, (d) GAHT, (e) MASSFormer, (f) S2Mamba, (g) MambaLG, (h) MambaHSI, (i) MambaMoE, and (j) PNEC-Mamba.}
	\label{fig:hanchuan_maps_appendix}
\end{figure*}

This appendix provides class-wise results and classification maps for all three datasets. Tables~\ref{tab:up_detailed}--\ref{tab:hanchuan_main_appendix} report mean class-wise accuracy and the mean $\pm$ standard deviation of OA, AA, and $\kappa$ over ten runs. Figures~\ref{fig:up_maps_appendix}--\ref{fig:hanchuan_maps_appendix} show the highest-OA run of each method with consistent class colors and background masks. Together, they supplement the aggregate results in the main paper with class-level and spatial comparisons.

\FloatBarrier

\begin{table*}[!t]
	\centering
	\tablefont
	\setlength{\tabcolsep}{4pt}
	\renewcommand{\arraystretch}{1.15}
	\begin{tabular}{@{}c||cc||cc||cccc||>{\columncolor{pneccol}}c@{}}
		\toprule
		& \multicolumn{2}{c||}{CNN-based} & \multicolumn{2}{c||}{Transformer-based} & \multicolumn{4}{c||}{Mamba-based} & \cellcolor{pneccol}{Proposed} \\
		\cmidrule(lr){2-3}\cmidrule(lr){4-5}\cmidrule(lr){6-9}\cmidrule(l){10-10}
		Class & DBMA & DBDA & GAHT & MASSFormer & S2Mamba & MambaLG & MambaHSI & MambaMoE & PNEC-Mamba \\
		\midrule
		1 & 87.37 & \textbf{96.11} & 89.01 & 93.96 & 93.47 & 90.50 & 89.44 & 94.45 & 94.74 \\
		2 & 92.41 & 92.96 & 82.21 & 93.82 & 90.77 & 93.55 & 93.16 & 92.05 & \textbf{97.74} \\
		3 & 88.02 & 92.98 & 56.62 & 92.57 & 92.75 & 91.40 & 94.00 & 94.99 & \textbf{98.67} \\
		4 & 92.93 & 94.06 & 94.94 & 92.07 & 95.35 & 91.33 & 78.49 & \textbf{96.06} & 88.42 \\
		5 & \textbf{100.00} & \textbf{100.00} & \textbf{100.00} & \textbf{100.00} & \textbf{100.00} & \textbf{100.00} & \textbf{100.00} & \textbf{100.00} & \textbf{100.00} \\
		6 & 99.01 & \textbf{99.84} & 93.85 & 98.41 & 97.90 & 97.60 & 99.08 & 97.73 & 99.33 \\
		7 & 98.38 & 99.16 & 83.76 & 99.61 & 98.37 & 87.32 & 94.85 & \textbf{99.83} & 98.68 \\
		8 & 86.06 & 85.86 & 83.81 & 89.51 & 94.07 & 96.68 & 96.53 & \textbf{99.14} & 98.66 \\
		9 & 92.47 & 95.60 & \textbf{97.35} & 86.30 & 85.42 & 94.54 & 93.09 & 96.31 & 86.31 \\
		\midrule
		OA & \res{92.10}{3.49} & \res{94.19}{1.32} & \res{85.36}{2.81} & \res{94.03}{0.73} & \res{93.14}{0.87} & \res{93.59}{0.87} & \res{92.82}{0.65} & \res{94.71}{1.41} & \bestres{96.77}{0.43} \\
		AA & \res{92.96}{2.86} & \res{95.17}{1.08} & \res{86.84}{3.93} & \res{94.03}{0.86} & \res{94.23}{0.59} & \res{93.66}{0.56} & \res{93.18}{0.48} & \bestres{96.73}{0.76} & \res{95.84}{0.55} \\
		$\kappa$ & \res{89.72}{4.47} & \res{92.41}{1.71} & \res{81.18}{3.41} & \res{92.17}{0.95} & \res{91.06}{1.11} & \res{91.59}{1.11} & \res{90.60}{0.83} & \res{93.09}{1.78} & \bestres{95.73}{0.56} \\
		\bottomrule
	\end{tabular}
	\caption{Detailed class-wise results on Pavia University. Class-wise entries are mean values; OA, AA, and $\kappa$ are reported as mean $\pm$ standard deviation.}
	\label{tab:up_detailed}
\end{table*}

\begin{table*}[!t]
	\centering
	\tablefont
	\setlength{\tabcolsep}{4pt}
	\renewcommand{\arraystretch}{1.15}
	\begin{tabular}{@{}c||cc||cc||cccc||>{\columncolor{pneccol}}c@{}}
		\toprule
		& \multicolumn{2}{c||}{CNN-based} & \multicolumn{2}{c||}{Transformer-based} & \multicolumn{4}{c||}{Mamba-based} & \cellcolor{pneccol}{Proposed} \\
		\cmidrule(lr){2-3}\cmidrule(lr){4-5}\cmidrule(lr){6-9}\cmidrule(l){10-10}
		Class & DBMA & DBDA & GAHT & MASSFormer & S2Mamba & MambaLG & MambaHSI & MambaMoE & PNEC-Mamba \\
		\midrule
		1 & 81.47 & 91.76 & 95.61 & 95.34 & 88.12 & 97.66 & 96.46 & 96.27 & \textbf{99.41} \\
		2 & 92.91 & 94.61 & 91.11 & 96.22 & 90.08 & 94.09 & \textbf{97.07} & 84.63 & 95.20 \\
		3 & 95.43 & 98.81 & 99.44 & 99.53 & 99.27 & 99.74 & \textbf{100.00} & 99.82 & \textbf{100.00} \\
		4 & 93.52 & 96.82 & 94.26 & \textbf{99.24} & 90.01 & 97.31 & 97.84 & 96.71 & 99.15 \\
		5 & 96.96 & 99.76 & \textbf{100.00} & 99.99 & \textbf{100.00} & 99.72 & 99.84 & \textbf{100.00} & \textbf{100.00} \\
		6 & 92.42 & 99.13 & 98.16 & 98.97 & 87.61 & 98.55 & 98.52 & 98.87 & \textbf{99.94} \\
		7 & 79.22 & 84.95 & 76.62 & 72.21 & 85.89 & 79.29 & 85.75 & 86.29 & \textbf{87.21} \\
		8 & 68.21 & 69.56 & 72.77 & 71.51 & 66.33 & 74.56 & 65.83 & 67.33 & \textbf{79.77} \\
		9 & 82.20 & 83.24 & 79.42 & 80.80 & 79.51 & 78.95 & 79.96 & 83.33 & \textbf{85.52} \\
		10 & 80.37 & 89.26 & 90.12 & 95.02 & 67.70 & 96.92 & 96.25 & 94.80 & \textbf{100.00} \\
		11 & 78.10 & 82.62 & 74.93 & \textbf{96.48} & 79.69 & 79.66 & 76.33 & 91.65 & 80.80 \\
		12 & 74.86 & 83.74 & 78.42 & \textbf{90.21} & 54.38 & 83.00 & 80.45 & 79.33 & 80.99 \\
		13 & 90.86 & 86.89 & \textbf{97.40} & 95.77 & 84.49 & 93.70 & 90.11 & 93.63 & 94.47 \\
		14 & 99.35 & 99.93 & 99.88 & \textbf{100.00} & \textbf{100.00} & 99.30 & 99.83 & \textbf{100.00} & \textbf{100.00} \\
		15 & 99.47 & \textbf{100.00} & \textbf{100.00} & \textbf{100.00} & \textbf{100.00} & 99.97 & 99.53 & \textbf{100.00} & 99.91 \\
		\midrule
		OA & \res{85.01}{5.08} & \res{89.25}{1.38} & \res{87.65}{0.68} & \res{91.23}{0.65} & \res{82.81}{2.08} & \res{89.86}{0.99} & \res{89.32}{0.51} & \res{89.82}{1.07} & \bestres{92.18}{0.38} \\
		AA & \res{87.02}{4.95} & \res{90.74}{1.15} & \res{89.88}{0.54} & \res{92.75}{0.62} & \res{84.87}{1.52} & \res{91.49}{0.82} & \res{90.92}{0.43} & \res{91.51}{0.83} & \bestres{93.49}{0.31} \\
		$\kappa$ & \res{83.80}{5.50} & \res{88.38}{1.50} & \res{86.66}{0.74} & \res{90.53}{0.70} & \res{81.44}{2.23} & \res{89.04}{1.07} & \res{88.46}{0.55} & \res{89.01}{1.15} & \bestres{91.55}{0.41} \\
		\bottomrule
	\end{tabular}
	\caption{Detailed class-wise results on Houston. Class-wise entries are mean values; OA, AA, and $\kappa$ are reported as mean $\pm$ standard deviation.}
	\label{tab:houston_detailed}
\end{table*}

\begin{table*}[!t]
	\centering
	\tablefont
	\setlength{\tabcolsep}{2.8pt}
	\renewcommand{\arraystretch}{1.05}
	
	\newcommand{\hcres}[2]{%
		\shortstack{$#1$\\$\pm #2$}%
	}
	\newcommand{\hcbestres}[2]{%
		\shortstack{$\mathbf{#1}$\\$\mathbf{\pm #2}$}%
	}
	
	\begin{tabular}{@{}c||cc||cc||cccc||>{\columncolor{pneccol}}c@{}}
		\toprule
		& \multicolumn{2}{c||}{CNN-based}
		& \multicolumn{2}{c||}{Transformer-based}
		& \multicolumn{4}{c||}{Mamba-based}
		& \cellcolor{pneccol}{Proposed} \\
		\cmidrule(lr){2-3}
		\cmidrule(lr){4-5}
		\cmidrule(lr){6-9}
		\cmidrule(l){10-10}
		Class & DBMA & DBDA & GAHT & MASSFormer & S2Mamba & MambaLG & MambaHSI & MambaMoE & PNEC-Mamba \\
		\midrule
		1  & 90.75 & 93.68 & 87.58 & 92.92 & 92.35 & 88.54 & 92.86 & 94.39 & \textbf{98.69} \\
		2  & 85.31 & 87.38 & 83.51 & 87.73 & 87.90 & 84.24 & 78.08 & 86.64 & \textbf{89.63} \\
		3  & 92.88 & 95.70 & 88.48 & 93.72 & 95.02 & 94.01 & 98.03 & 92.02 & \textbf{99.43} \\
		4  & 96.79 & 97.61 & 96.39 & 98.90 & 97.31 & 96.86 & 98.94 & 94.03 & \textbf{99.84} \\
		5  & \textbf{99.85} & 99.74 & 98.32 & 99.60 & 99.55 & 99.85 & 99.70 & 99.32 & 99.68 \\
		6  & 82.53 & \textbf{86.94} & 73.46 & 83.40 & 83.48 & 80.30 & 62.55 & 61.75 & 32.11 \\
		7  & 95.11 & 96.65 & 89.98 & 94.67 & 93.86 & 92.18 & 92.10 & 78.50 & \textbf{97.66} \\
		8  & 74.27 & 73.67 & 71.34 & 75.27 & \textbf{79.94} & 77.87 & 56.97 & 78.93 & 77.49 \\
		9  & 92.45 & \textbf{96.51} & 87.85 & 90.73 & 92.35 & 89.74 & 91.67 & 87.34 & 92.65 \\
		10 & 96.17 & 96.62 & 97.32 & 96.95 & \textbf{98.06} & 95.99 & 92.88 & 93.67 & 94.69 \\
		11 & 94.43 & 96.47 & 90.53 & 96.43 & 96.86 & 94.36 & 91.47 & 93.49 & \textbf{97.94} \\
		12 & 92.48 & 88.98 & 85.23 & 90.54 & 84.79 & \textbf{92.66} & 82.35 & 85.24 & 84.59 \\
		13 & 81.98 & \textbf{88.11} & 83.85 & 80.61 & 83.11 & 79.76 & 84.43 & 82.98 & 86.16 \\
		14 & 82.76 & 86.62 & 82.17 & 86.57 & 83.39 & 87.46 & 90.35 & 89.32 & \textbf{95.60} \\
		15 & 99.60 & \textbf{99.93} & 99.10 & 97.68 & 99.87 & 98.54 & 97.91 & 99.30 & 96.34 \\
		16 & 98.14 & 98.28 & 98.46 & 98.55 & 97.68 & 97.68 & 98.13 & 93.16 & \textbf{99.17} \\
		\midrule
		OA
		& \hcres{91.20}{0.87}
		& \hcres{92.87}{1.00}
		& \hcres{89.45}{0.89}
		& \hcres{92.35}{0.37}
		& \hcres{92.26}{0.27}
		& \hcres{90.92}{1.47}
		& \hcres{89.67}{1.14}
		& \hcres{90.03}{3.61}
		& \hcbestres{94.11}{0.43} \\
		AA
		& \hcres{90.97}{0.77}
		& \hcbestres{92.68}{1.33}
		& \hcres{88.35}{1.02}
		& \hcres{91.52}{0.48}
		& \hcres{91.60}{0.23}
		& \hcres{90.63}{1.03}
		& \hcres{88.03}{0.98}
		& \hcres{88.13}{3.89}
		& \hcres{90.10}{1.16} \\
		$\kappa$
		& \hcres{89.74}{1.00}
		& \hcres{91.69}{1.16}
		& \hcres{87.71}{1.04}
		& \hcres{91.07}{0.43}
		& \hcres{90.97}{0.31}
		& \hcres{89.43}{1.69}
		& \hcres{87.93}{1.33}
		& \hcres{88.42}{4.11}
		& \hcbestres{93.10}{0.50} \\
		\bottomrule
	\end{tabular}
	\caption{Classification results on WHU-Hi-HanChuan. Class-wise entries are mean values; OA, AA, and $\kappa$ are reported as mean $\pm$ standard deviation.}
	\label{tab:hanchuan_main_appendix}
	\vspace{8pt}
	\refstepcounter{algorithm}
	\noindent\textbf{Algorithm~\thealgorithm: Training and Inference Procedure of PNEC-Mamba}
	\label{alg:pnec}
	\par\vspace{2pt}\hrule\vspace{2pt}
	\small
	\begin{algorithmic}[1]
		\REQUIRE Hyperspectral image $X$, labeled map $Y$, model parameters $\theta$, prototype bank $\mathcal{P}$, total epochs $T$, and warm-up epochs $T_w$
		\ENSURE Full-resolution prediction $\hat{Y}$
		\STATE Initialize $\theta$ and $\mathcal{P}$
		\FOR{$t=1$ to $T$}
		\STATE Extract complementary representations $F^{\mathrm{spa}}$ and $F^{\mathrm{diff}}$, and fuse them to obtain $F$
		\STATE Update $\mathcal{P}$ using labeled representations and momentum
		\STATE Compute pixel--prototype similarities and prototype probabilities
		\STATE Construct positive evidence $E^{+}$ and negative evidence $E^{-}$
		\STATE Estimate multi-source uncertainty and obtain the calibration gate $G$
		\IF{$t \leq T_w$}
		\STATE Set the calibration strength $s \leftarrow 0$
		\ELSE
		\STATE Set the calibration strength $s \leftarrow 1$
		\ENDIF
		\STATE Calibrate $F^{\mathrm{cal}} \leftarrow F+sG(\alpha E^{+}-\beta E^{-})$ and predict logits $Z^{\mathrm{cal}}$
		\STATE Compute $\mathcal{L} \leftarrow \mathcal{L}_{\mathrm{cls}}+\lambda_{\mathrm{br}}\mathcal{L}_{\mathrm{br}}+\lambda_{\mathrm{pro}}\mathcal{L}_{\mathrm{pro}}+\lambda_{\mathrm{con}}\mathcal{L}_{\mathrm{con}}$
		\STATE Update $\theta$ by back-propagation
		\ENDFOR
		\STATE Obtain $Z^{\mathrm{cal}}$ from the trained model with $s=1$
		\STATE Compute $P^{\mathrm{up}} \leftarrow \operatorname{Softmax}(\operatorname{Up}(Z^{\mathrm{cal}}))$
		\STATE Compute $\bar{P}^{\mathrm{up}} \leftarrow \operatorname{LocalAvg}(P^{\mathrm{up}})$
		\STATE Recover $P^{\mathrm{rec}} \leftarrow (1-\eta)P^{\mathrm{up}}+\eta\bar{P}^{\mathrm{up}}$
		\STATE Obtain $\hat{Y}\leftarrow\operatorname{Argmax}(P^{\mathrm{rec}})$
		\RETURN $\hat{Y}$
	\end{algorithmic}
	\vspace{2pt}\hrule
\end{table*}

\end{document}